\documentclass[runningheads]{llncs}
\usepackage{graphicx}

\usepackage{tikz}
\usepackage{comment}
\usepackage{amsmath,amssymb} 
\usepackage{color}

\usepackage{epsfig}

\usepackage{multirow}
\usepackage{xcolor}
\usepackage{hyperref}

\DeclareMathOperator*{\argminB}{argmin}

\usepackage[accsupp]{axessibility}  

\begin{document}
\pagestyle{headings}
\mainmatter
\def\ECCVSubNumber{7955}  

\title{DSR -- A dual subspace re-projection network for surface anomaly detection}

\titlerunning{DSR -- A dual subspace re-projection network for surface anomaly detection}

\author{Vitjan Zavrtanik \and Matej Kristan \and Danijel Skočaj}
\institute{University of Ljubljana, Faculty of Computer and Information Science\\
\email{\{vitjan.zavrtanik, matej.kristan, danijel.skocaj\}@fri.uni-lj.si}}

\authorrunning{Zavrtanik et al.}
\maketitle

\begin{abstract}

The state-of-the-art in discriminative unsupervised surface anomaly detection relies on external datasets for synthesizing anomaly-augmented training images. Such approaches are prone to failure on near-in-distribution anomalies since these are difficult to be synthesized realistically due to their similarity to anomaly-free regions. We propose an architecture based on quantized feature space representation with dual decoders, DSR, that avoids the image-level anomaly synthesis requirement. Without making any assumptions about the visual properties of anomalies, DSR generates the anomalies at the feature level by sampling the learned quantized feature space, which allows a controlled generation of near-in-distribution anomalies.
DSR achieves state-of-the-art results on the KSDD2 and MVTec anomaly detection datasets. The experiments on the challenging real-world KSDD2 dataset show that DSR significantly outperforms other unsupervised surface anomaly detection methods, improving the previous top-performing methods by $10\%$ AP in anomaly detection and $35\%$ AP in anomaly localization. Code is available at: \href{https://github.com/VitjanZ/DSR_anomaly_detection}{https://github.com/VitjanZ/DSR\_anomaly\_detection}.
\end{abstract}

\section{Introduction}

Surface anomaly detection addresses localization of image regions that deviate from normal object appearance. This is a fundamental problem in industrial inspection, in which the anomalies are defects on production line objects. In the most challenging situations, the distribution of the normal appearance of the inspected objects is very close to the distribution of anomaly appearances, while anomalies often occupy only a small portion of the object. Furthermore, the anomalies are rare in practical production lines, making the acquisition of a suitable data set for training supervised methods infeasible. The methods thus focus on leveraging only anomaly-free images, since these can be abundantly obtained.

\begin{figure}
\centering
  \includegraphics[width=0.7
  \linewidth]{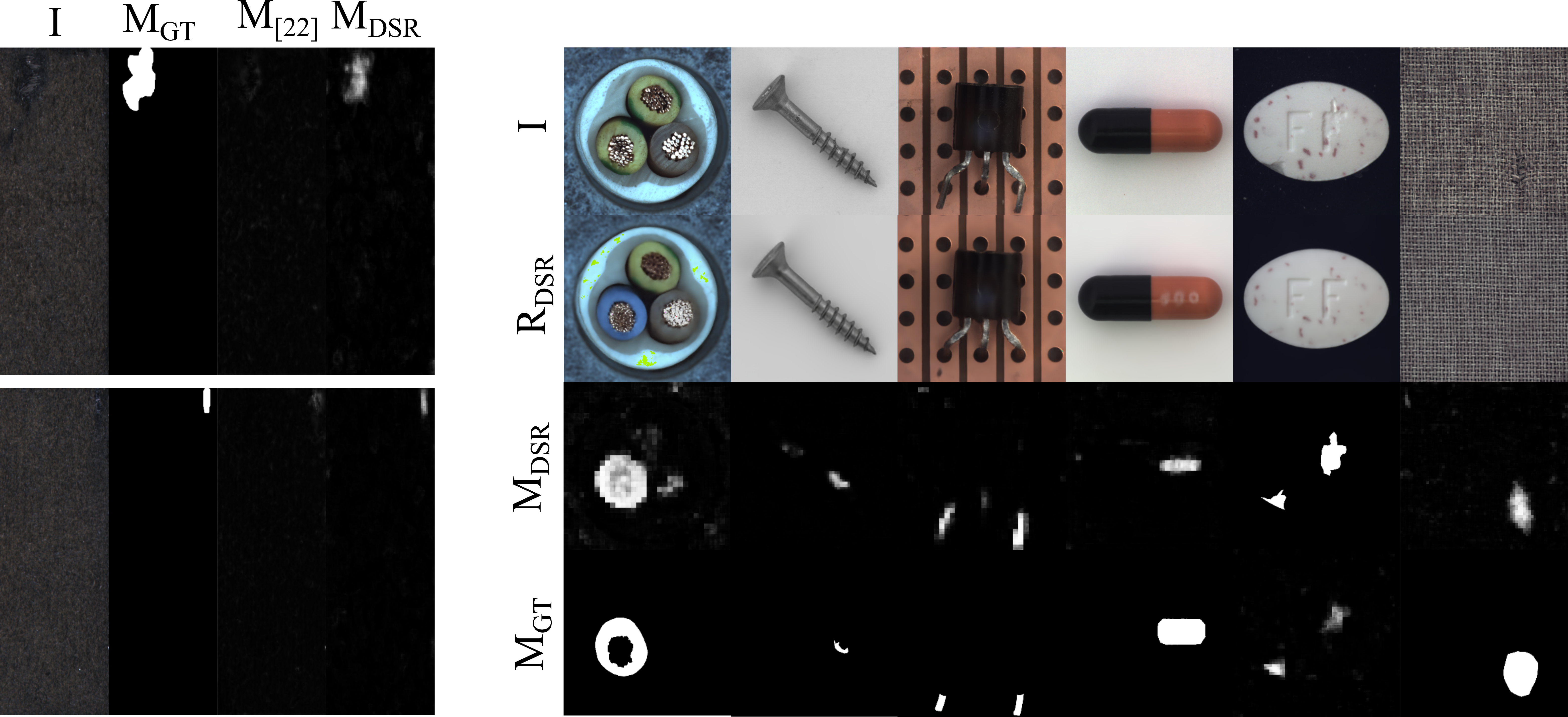}
\caption{
The dual decoder architecture with discrete feature space allows DSR robust object-specific reconstruction ($R_{DSR}$) and accurate detection of near-in-distribution anomalies, which present a considerable challenge for the recent state-of-the-art ($M_{[22]}$).
}
\label{fig:first}
\end{figure}

Most anomaly detection approaches are based on computing the difference between the inspected image and its image-level or feature-level reconstruction~\cite{akcay2018ganomaly,akccay2019skip,schlegl2019f,bergmann2018improving,zavrtanik2020riad,bergmann2019mvtec}, with their reconstruction method trained only on anomaly-free images. These approaches assume that anomalies will be poorly reconstructed since they have never been observed during training and that the reconstruction failure on anomalies can be well detected by $L_2$ or SSIM~\cite{wang2004image} difference with the input image.

However, $L_2$ and SSIM measures can only detect anomalies that differ substantially from normal appearance.
Subsequent works have addressed this problem by either learning the distance measure with a discriminative network~\cite{zavrtanik2021draem} or by classifying the anomalies directly on the input image~\cite{li2021cutpaste}. These methods require annotated anomalies at training time, and resort to simulation of anomalies from auxiliary datasets and copy-pasting and blending them with the anomaly-free training images.
While these methods by far outperform the reconstruction-only methods, they rely substantially on the quality of the auxiliary dataset and the simulation process quality; their performance still degrades on near-in-distribution anomalies (Figure~\ref{fig:first}) since it is difficult to simulate these realistically.

In this paper we address two drawbacks of the surface anomaly detection state-of-the-art: the reliance on the auxiliary anomaly simulation datasets and poor near-in-distribution anomaly detection.
We propose a dual subspace re-projection surface anomaly detection network (DSR). The network leverages the 
framework of discretized latent feature space image representation~\cite{vqvae2}, and jointly learns a general and a normal-appearance-specific subspace re-projection to emphasize the anomaly detection capability. 

The proposed architecture avoids reliance on auxiliary anomaly datasets in training anomaly discrimination. 
We propose a new anomaly simulation technique that generates the anomalies directly from the network's discretized latent space of natural images, leading to significant performance improvements on near-in-distribution anomaly detection (Figure~\ref{fig:first}). 

Our contribution is thus twofold: (i) the dual image reconstruction branch architecture with discretized latent representation and (ii) the latent space anomaly generation method that leverages the learned representation of natural images. DSR substantially outperforms the state-of-the-art in near-in-distribution anomaly detection on the recent challenging KSDD2~\cite{bovzivc2021mixed} and delivers state-of-the-art performance on the standard MVTec anomaly detection dataset~\cite{bergmann2019mvtec}.

\section{Related Work}
Many recent surface anomaly detection methods are based on image reconstruction \cite{akcay2018ganomaly,akccay2019skip,bergmann2018improving}, where an encoder-decoder network is trained for image reconstruction on anomaly free images. The anomaly detection capability of these methods is based on the assumption that the trained networks will be unable to accurately reconstruct anomalous regions due to never seeing them during training, making anomalies detectable by comparing the input image to its reconstruction. The core assumption often does not hold, especially in more diverse datasets, as the reconstruction networks learn to generalize well which enables accurate anomaly reconstruction, hampering the downstream anomaly detection performance. Similarly, in \cite{zavrtanik2020riad} iterative inpainting is used for image reconstruction, however, the method is sensitive to random pattern regions that are difficult to inpaint and which cause false positive detections.

Several recent approaches utilize the ability of pretrained networks to extract informative features from an image. In~\cite{bergmann2020uninformed} a network is trained on anomaly free images to reconstruct features extracted by the pretrained network. In~\cite{featspace1} a feature map is generated as a concatenation of several layers of a pretrained network. An auto-encoder is then trained to reconstruct the resulting feature map. As with image reconstruction based anomaly detection methods, the networks in \cite{bergmann2020uninformed,featspace1} are assumed to be unable to reconstruct the features extracted by the pretrained networks. In~\cite{featspace2}, a pretrained network is also used to extract informative features from anomaly free data, however instead of relying on reconstruction for anomaly detection, a multivariate gaussian is fit to the anomaly-free data. A Mahalanobis distance is then used as an anomaly score. A similar approach is used in \cite{defard2021padim}, however, a Gaussian is fit at each location in the feature map.

Discriminative unsupervised anomaly detection methods~\cite{zavrtanik2021draem,disc1,li2021cutpaste} utilize synthetically generated anomalies to train a discriminative anomaly detection network.  In order to alleviate overfitting on the synthetic anomaly appearance, in \cite{zavrtanik2021draem} a reconstruction network is used to restore the normal appearance of the synthetic anomalies. The discriminative network then learns a distance function between the original image and its reconstruction to perform anomaly detection. Due to a limited distribution of the generated synthetic anomalies, the reconstruction network may overfit to the synthetic appearance and fail to restore normality in near-distribution anomalies, leading to a reduction in performance in downstream anomaly detection.

\begin{figure*}[!htb]
\centering
  \includegraphics[width=1.0\linewidth]{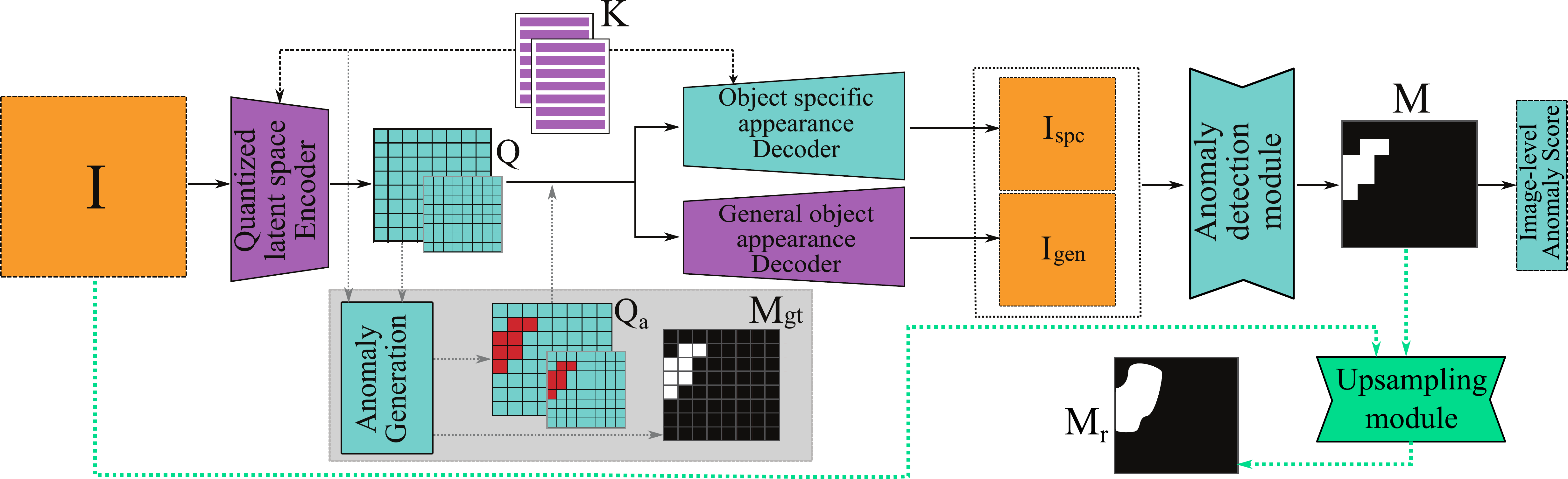}
\caption{
The DSR architecture. During training, the non-anomalous image quantized feature maps $(Q_{hi},Q_{lo})$ are replaced by the anomaly augmented feature maps $(Q_{a,hi},Q_{a,lo})$ generated by the latent space sampling procedure (shaded block). The pathway marked with green arrows are used when training the Upsampling module with simulated smudges and at inference.
}
\label{fig:total_architecture}
\end{figure*}

\section{DSR}

The DSR architecture is outlined in Figure~\ref{fig:total_architecture}. The input image is projected into a quantized latent space (Section \ref{sec:discreteEncoder}) and then decoded in parallel by two decoders specialized on different appearance subspaces. One decoder, the general object appearance decoder (Section~\ref{sec:discreteDecoder}), is trained for high-fidelity reconstruction of arbitrary natural images, while the second, object-specific decoder (Section~\ref{sec:objectSpecific}), is restricted to reconstructing only normal local appearances of the selected object. The two reconstructed images are then analyzed by the anomaly detection module (Section \ref{sec:discriminative}). The output localization map is at feature resolution and is
upsampled to the input image resolution by the Upsampling module (Section \ref{sec:upsampling}).

DSR is trained by a novel technique of sampling the quantized latent feature space for generating near-in-distribution anomalies (Section~\ref{sec:anomgen}). The DSR training regime is explained in Section~\ref{sec:train}.

\subsection{Quantized latent space encoder}
\label{sec:discreteEncoder}

DSR leverages quantized feature space representation, which has recently demonstrated strong modelling capabilities of complex natural image distributions
,e.g., \cite{Esser_2021_CVPR,ramesh2021zero}. The approach is based on quantizing the extracted features with features from a codebook $\mathbf{K}$ which has been trained for optimal decoding of spatial configurations of quantized features into high-fidelity images.

\begin{figure}
\centering
  \includegraphics[width=0.9\linewidth]{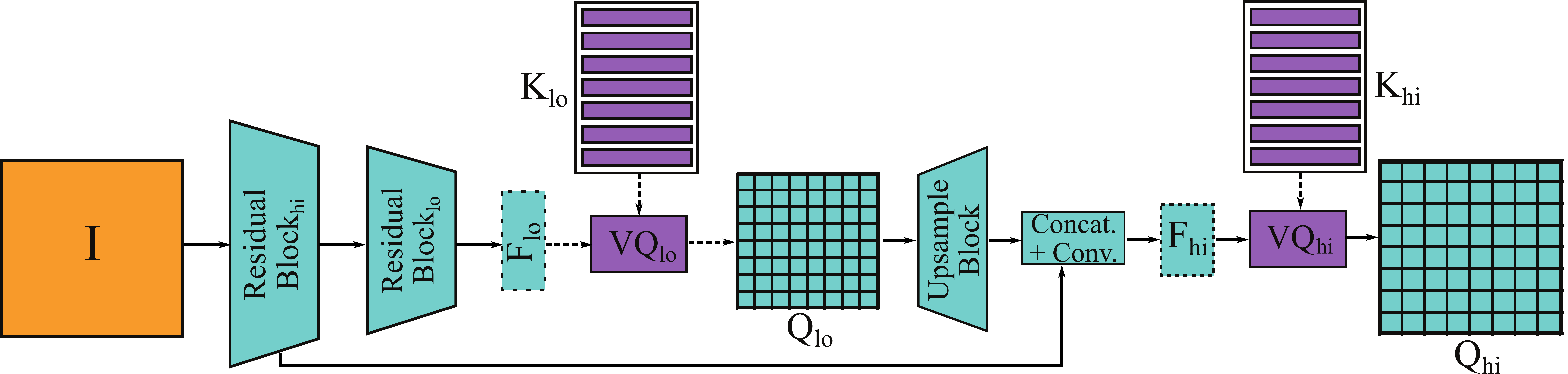}
\caption{The quantized latent space encoder architecture. Two residual blocks extract image features at different spatial resolutions. The low-resolution features $F_{lo}$ are quantized by the codebook $K_{lo}$ into a feature representation $Q_{lo}$. The high-resolution features are concatenated by upsampled $Q_{lo}$, followed by a convolutional block and quantized by the codebook $K_{hi}$ into a high-resolution representation  $Q_{hi}$.}
\label{fig:latentenc}
\end{figure}

In particular, the quantized latent space encoder module in Figure~\ref{fig:total_architecture} accepts the input image $\mathbf{I}$ and projects it into a feature space $\mathbf{F}$ using a Resnet-based encoder. A quantized feature representation $\mathbf{Q}$ of the input image is obtained by replacing each feature vector $\mathbf{F}_{ij}$ with its nearest neighbor $\mathbf{e}_l$ in $\mathbf{K}$, i.e,
\begin{equation}
    Q_{i,j} = q(F_{i,j}) = \argminB_{\mathbf{e}_l \in \mathbf{K}}  \big( ||\mathbf{F}_{i,j} - \mathbf{e}_l || \big).
\end{equation}
In the following, we refer to this operation as vector quantization (VQ). 
Note that the input image is encoded at two levels of detail using low- and high-resolution codebooks ($\mathbf{K}_\mathrm{lo}$, $\mathbf{K}_\mathrm{hi}$), producing $\mathbf{Q}_\mathrm{lo}$ and $\mathbf{Q}_\mathrm{hi}$. The two-level VQ has recently been reported to produce superior reconstructions~\cite{vqvae2}. The architecture of the quantized latent space encoder is shown in Figure \ref{fig:latentenc}. The quantized feature maps $Q_{hi}$ and $Q_{low}$ produced by the quantized latent space encoder are $4\times$ and $8\times$ smaller than the original input image, respectively.

\subsection{General object appearance decoder} \label{sec:discreteDecoder}

The subspace of VQ encoded natural images is captured by specific \textit{spatial configurations} of quantized feature vectors. We apply a \textit{general object appearance decoder} 
to learn decoding of these configurations into image reconstructions. The decoder first upsamples the low resolution $\mathbf{Q}_\mathrm{lo}$ and concatenates it with the $\mathbf{Q}_\mathrm{hi}$, which is followed by two ResNet blocks and two transposed convolution upsampling blocks that map into the reconstructed image $\mathbf{I}_\mathrm{gen}$.

\subsection{Object-specific appearance decoder} \label{sec:objectSpecific}

\begin{figure}
\centering
  \includegraphics[width=0.9\linewidth]{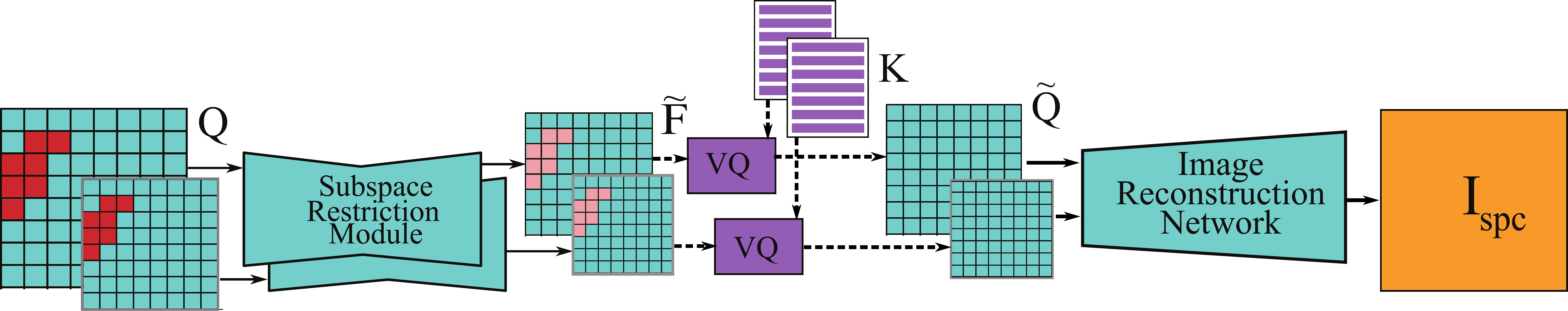}
\caption{
The object-specific appearance decoder architecture. 
The features in the quantized features maps $\mathbf{Q}$ are reduced by a subspace restriction module into non-quantized features $\tilde{\mathbf{F}}$ that are then vector-quantized (VQ) by the codebooks $\mathbf{K}$ into $\tilde{\mathbf{Q}}$, which are then decoded into an anomaly-free image $\mathbf{I}_{spc}$.
}
\label{fig:reconstructive}
\end{figure}

\begin{figure}
\centering
  \includegraphics[width=1.0\linewidth]{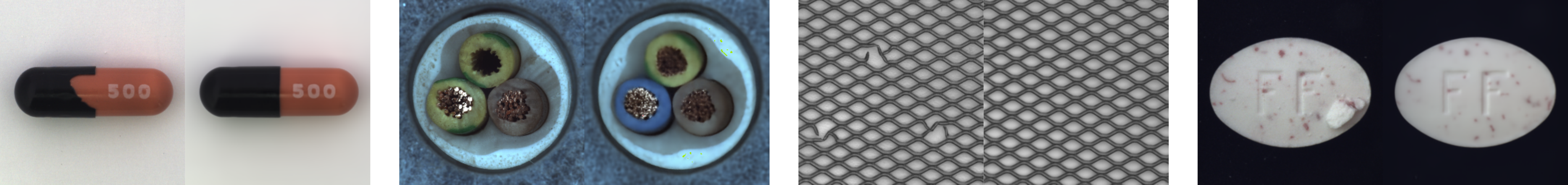}
\caption{
The object-specific decoder reconstructs the anomalous images (left) into anomaly-free images (right) with remarkable fidelity.
In the second example, it even correctly reconstructs the anomalous green wire into blue.
}
\label{fig:objspecrecon}
\end{figure}

The tasks of the second decoder, the \textit{object-specific appearance decoder} (see Figure~\ref{fig:reconstructive}) is to restore local visual anomalies into feasible normal appearances of the object instances observed during training. In particular, we would like to restrict the appearance subspace, i.e., the allowed spatial VQ feature configurations, into configurations that agree with normal appearances. This is achieved by a \textit{subspace restriction module} (Figure~\ref{fig:reconstructive}), which transforms both high- and low-resolution general input VQ representations ($\mathbf{Q}$=$\{\mathbf{Q}_\mathrm{hi}$, $\mathbf{Q}_\mathrm{lo}\}$) into non-quantized object-specific subspace configurations ($\tilde{\mathbf{F}}$=$\{\tilde{\mathbf{F}}_\mathrm{lo}$, $\tilde{\mathbf{F}}_\mathrm{hi}\}$). This is followed by a VQ projection (with codebooks $\mathbf{K}$=\{$K_{hi},K_{lo}\}$) into object subspace-restricted quantized feature configurations ($\tilde{\mathbf{Q}}$=$\{\tilde{\mathbf{Q}}_\mathrm{hi}$, $\tilde{\mathbf{Q}}_\mathrm{lo}\}$). The quantized representations are concatenated (the low-resolution representation is upsampled first) and decoded into reconstructed anomaly-free image $\mathbf{I}_\mathrm{spc}$ using a convolutional decoder. The subspace restriction modules (Figure \ref{fig:reconstructive}) are encoder-decoder convolutional networks with three downsampling and corresponding upsampling blocks. The image reconstruction network consists of two downsampling convolutional blocks, followed by four transposed convolution upsampling blocks.
Examples of images with anomalies present and their anomaly-free reconstructions are shown in Figure~\ref{fig:objspecrecon}.

\subsection{Anomaly detection module} \label{sec:discriminative}

The purpose of the anomaly detection module is to localize the anomaly by inspecting the input image reconstruction generated by the general object appearance decoder ($\mathbf{I}_\mathrm{gen}$) and the object-specific appearance decoder ($\mathbf{I}_\mathrm{spc}$). The reconstructed images are concatenated depth-wise and decoded into a segmentation mask $\mathbf{M}$ by a Unet-based architecture. $\mathbf{M}$ is the output anomaly map indicating the pixel-level location of the anomalies in the image. To compute also the image-level anomaly score, we apply a simple segmentation mask interpretation procedure as in~\cite{zavrtanik2020riad} -- the segmentation mask is smoothed by a $21 \times 21$ averaging filter and globally max-pooled into a single score.

\subsection{Upsampling module} \label{sec:upsampling}

The generated segmentation mask  $\mathbf{M}$ is of the same resolution as the feature maps.
A simple Unet-like upsampling module is thus used to resample the mask to full resolution. The input to the network is a depth-wise concatenation of the input image  $\mathbf{I}$ and bilinearly upsampled mask  $\mathbf{M}$. The output is the final full-resolution mask  $\mathbf{M_r}$ (Figure~\ref{fig:total_architecture}).

\subsection{Feature-space surface anomaly generation} \label{sec:anomgen}

The purpose of anomaly generator is to simulate near-in-distribution anomalies of various shapes and sizes with diverse visual appearances to (i) learn normal appearance subspace restriction in the 
object-specific appearance decoder (Section~\ref{sec:objectSpecific}) and (ii) to specialize the anomaly detection module (Section~\ref{sec:discriminative}) for detection of potentially gentle appearance deviations of anomalies from a diverse within-class normal appearance.

We propose a method that leverages the learned quantized subspace in DSR to generate such training anomalies as follows.
An anomaly-free input image $\mathbf{I}$ is encoded into a VQ subspace representation $\mathbf{Q}$ and an anomaly mask $\mathbf{M}_\mathrm{gt}$ is generated by sampling a Perlin noise~\cite{perlin1985image}, with values 1 indicating anomalous pixels. The features in $\mathbf{Q}$ corresponding to the anomaly indicators in $\mathbf{M}_\mathrm{gt}$ are replaced by sampling from the set of codebook features $\mathbf{K}$. 

Sampling without constraints will likely lead to significant appearance changes on anomalous pixels, resulting in trivial out-of-distribution reconstructed images. On the other hand, if the closest vectors are sampled that are too similar to the normal appearance, they will likely lead to false-positive detections.

We thus first define a similarity bound on all features for a given image $n \sim \mathcal{U}[\lambda_s N_K, N_K]$, where $N_K$ is the number of codebook vectors and $\lambda_s$ is the similarity bound parameter. Then we replace each feature in $\mathbf{Q}$, indicated by $\mathbf{M}_\mathrm{gt}$, with one of its near neighbors from the codebook feature vectors, sampled uniformly, i.e., $k \sim \mathcal{U}[\lambda_s N_K, n]$. In all experiments $\lambda_s$ is $0.05$, therefore, excluding the $5\%$ of most similar vectors to prevent false-positive generation, while prioritizing the selection of the features close to the vector to be replaced to encourage the generation of near-in-distribution anomalies.

\subsection{DSR training procedure}\label{sec:train}

The DSR is trained in three stages. In the \textbf{first stage},
the quantized latent space encoder (Section~\ref{sec:discreteEncoder}), along with the VQ codebook and the general object appearance decoder are trained on ImageNet~\cite{deng2009imagenet} to learn the subspace of natural images, which allows a high-fidelity reconstruction of general images. We use the procedure from~\cite{vqvae} that minimizes the image reconstruction loss as well as the difference between the feature space projection $\mathbf{F}$ computed in the quantized latent space encoder (Figures~\ref{fig:total_architecture} and \ref{fig:latentenc}) and its quantized version $\mathbf{Q}$, i.e,
\begin{equation}
    \mathcal{L}_\mathrm{st1} =  L_2(\mathbf{I}, \mathbf{I}_\mathrm{gen}) + L_2(sg[\mathbf{F}], \mathbf{Q}) + \lambda_1 L_2(\mathbf{F}, sg[\mathbf{Q}]),
\end{equation}
where $L_2(\cdot)$ is the Euclidean distance and $sg[\cdot]$ is the stop gradient operator constraining the operand to be a non-updated constant~\cite{vqvae}. 
After training, the discrete latent space encoder, the codebook and the general object appearance decoder (coloured in magenta in Figure~\ref{fig:total_architecture}) are fixed.

In the second, \textbf{anomaly detection training stage}, the detection parts of DSR are trained on images of the selected object type.
Anomaly-free training images are projected through the quantized latent space encoder into their quantized feature representation $\mathbf{Q}$. The surface anomaly generation method presented in Section~\ref{sec:anomgen} then generates the anomalies at the feature level, $\mathbf{Q}_\mathrm{a}$, along with their ground truth masks $\mathbf{M}_\mathrm{gt}$. The representation  $\mathbf{Q}_\mathrm{a}$ (that replaces $\mathbf{Q}$ in Figure~\ref{fig:total_architecture}) is then forward passed and decoded into the anomaly mask $\mathbf{M}$. 

The anomaly detection module and the object-specific appearance decoder are trained by minimizing the focal loss between the ground truth $\mathbf{M}_\mathrm{gt}$ and predicted anomaly mask $\mathbf{M}$, the $L_2$ distance between the subspace-restricted configurations $\tilde{\mathbf{F}}$ computed in the object-specific appearance decoder (Figure~\ref{fig:reconstructive}) and the non-anomalous input image quantized representation $\mathbf{Q}$, and the $L_2$ distance between the non-anomalous input image $\mathbf{I}$ and its object-specific reconstruction $\mathbf{I}_\mathrm{spc}$, i.e.,
\begin{equation}
    \mathcal{L}_\mathrm{st2} = \mathcal{L}_\mathrm{foc}(\mathbf{M}_\mathrm{gt}, \mathbf{M}) + \lambda_2 L_2(\tilde{\mathbf{F}},\mathbf{Q}) + \lambda_3 L_2(\mathbf{I},\mathbf{I}_\mathrm{spc}).  
\label{eq:reconloss}
\end{equation}

Finally, the Upsampling module is trained. After the anomaly detection network has been trained, images with copy-pasted smudges are generated and the low-resolution anomaly masks are computed as detection network outputs. Since the full-resolution masks are known from the smudge pasting on the original image, the upsampling network can be trained with a focal loss.

\section{Experiments}

\subsection{Implementation details}

In the first training stage (Section \ref{sec:train}) quantized latent space encoder, general object appearance decoder and the latent vector codebook are trained for image reconstruction on ImageNet~\cite{deng2009imagenet} for $200,000$ iterations with a batch size of 32 and a learning rate of $0.0002$. The codebooks $K_{hi}$ and $K_{lo}$ each contain $4096$ latent vectors of dimension $128$. In the second training stage (Section \ref{sec:train}) the object-specific appearance decoder and the anomaly detection model are trained for $100,000$ iterations with a batch size of $8$ with a learning rate of $0.0002$. The learning rate is decreased by a factor of $10$ after $80,000$ iterations. The $\lambda_2$ and $\lambda_3$  values in Eq.\ref{eq:reconloss} are set to $1$ and $10$, respectively. In the third stage the Upsampling module is then trained for $20,000$ iterations with a learning rate of $0.0002$ and a batch size of $8$. The training hyperparameters and network architectures remain constant throughout our experiments.

\subsection{Experimental results on KSDD2}

\begin{figure}
\centering
  \includegraphics[width=1.0\linewidth]{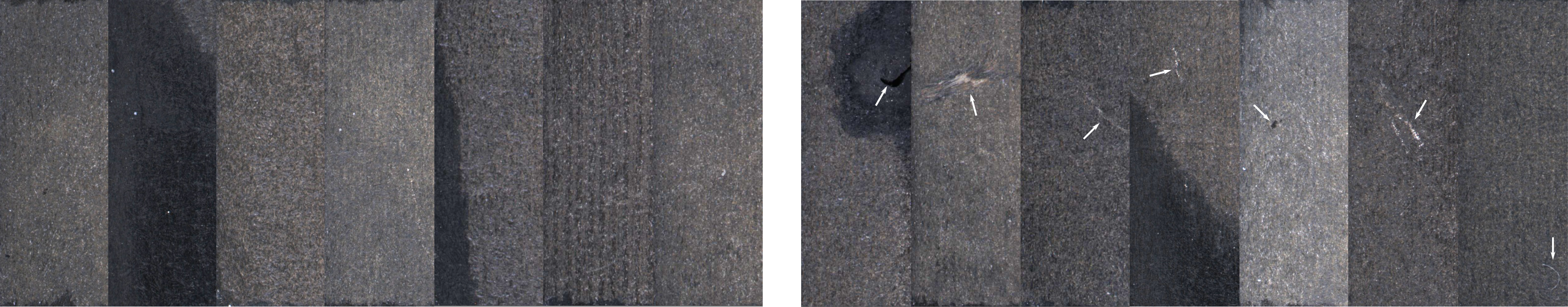}
\caption{
Normal (left) and anomalous (right) images from KSDD2. The anomalies (marked with white arrows) are very similar to the normal appearance in this dataset.}
\label{fig:ksdd2}
\end{figure}

The recently proposed KSDD2~\cite{bovzivc2021mixed} surface anomaly detection dataset is currently the most challenging dataset with near-in-distribution surface anomalies (see Figure~\ref{fig:ksdd2}). KSDD2 was acquired on real industrial production lines and contains a wide variety of anomalies, many of which are particularly challenging due to their similarity to the normal appearances in the training set.
 
The dataset contains $2085$ anomaly-free and $246$ anomalous training images to test anomaly detection under unsupervised and supervised setups. For example, in the unsupervised setup, only anomaly-free training images may be used. The test set contains $894$ anomaly-free and $110$ anomalous images. We follow the evaluation procedure defined in~\cite{bovzivc2021mixed}, with the AP metric for image-level anomaly detection ($AP_{det}$), and present also anomaly localization results in terms of pixel-level AP ($AP_{loc}$). 

Additional anomalous training samples available in KSDD2~\cite{bovzivc2021mixed} enable comparison of DSR with the supervised defect detection methods. Note that, although the acquisition of anomalous examples is difficult in practice, in many cases these are available, albeit in small quantities. Most of unsupervised anomaly detection methods, however, do not, or even can not, make use of this additional information, if available. The proposed DSR method can be easily adapted to take such annotations of anomalous images into account. In addition to the generated synthetic anomalies, the real-world anomalies with known ground truth can be utilised to train the anomaly detection module.

\begin{figure}
\centering
  \includegraphics[width=1.0\linewidth]{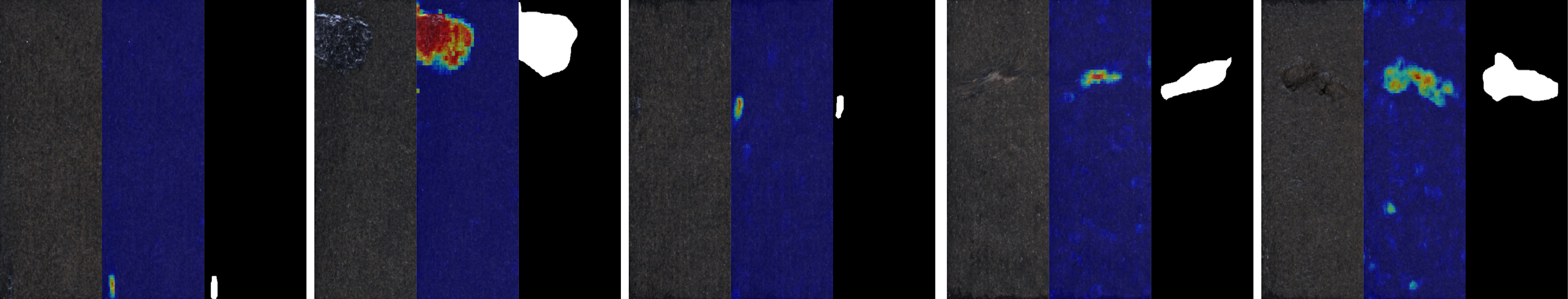}
\caption{Qualitative results of the unsupervised DSR on the KSDD2 dataset: the input image, the overlaid predicted mask and the ground truth.
}
\label{fig:ksdd2_DSR_ex}
\end{figure}

\subsubsection{Comparison with the state-of-the-art}

The results of unsupervised surface anomaly detection methods are shown in Table \ref{tab:ksdd2}.
DSR significantly outperforms other state-of-the-art unsupervised methods such as MAD~\cite{featspace2}, PaDim~\cite{defard2021padim} and DRAEM~\cite{zavrtanik2021draem}, achieving higher AP scores for anomaly detection and localization. It outperforms the previous best image-level AP score by $7.9$ p.p. Qualitative examples of the unsupervised DSR are presented in Figure \ref{fig:ksdd2_DSR_ex} and show that despite a quite heterogeneous normal appearance, that can also be observed in Figure~\ref{fig:ksdd2}, the visual defects are successfully detected and localized.

\subsubsection{Extension to supervised learning}

In contrast to most of the unsupervised visual anomaly detection methods, DSR can also utilise pixel-level anotations, if they are available, making it applicable in low-shot anomaly detection scenarios. We evaluated the proposed approach in the supervised setting. Table~\ref{tab:ksdd2ms} shows the comparison with the method from \cite{bovzivc2021mixed} that is specifically designed for supervised defect detection and can operate with various levels of supervision. In KSDD2, there are 246 anomalous samples available, and if image-level labels are available but none of them is segmented, the method~\cite{bovzivc2021mixed} operates in the weakly supervised mode ($N=0$). 

\begin{table}
\centering
\resizebox{0.6 \linewidth}{!}{\begin{tabular}{c c c c c c}
\hline
Method & US~\cite{bergmann2020uninformed} & MAD~\cite{featspace2} & DR$\AE$M~\cite{zavrtanik2021draem} & PaDim~\cite{defard2021padim} & DSR \\ \hline
$AP_{det}$ & 65.3 & 79.3 & 77.8 & 55.6 & \textbf{87.2} \\
$AP_{loc}$ & - & - & 42.4 & 45.3 & \textbf{61.4} \\ \hline
\end{tabular}}
\caption{Anomaly detection ($AP_{det}$) and localization ($AP_{loc}$) on the KSDD2 dataset.}
\label{tab:ksdd2}
\end{table}

The completely unsupervised DSR, without taking these additional positive training samples into account, outperforms the weakly supervised method~\cite{bovzivc2021mixed} for 13.9 p.p. in $AP_{det}$ and achieves a good localization result, whereas \cite{bovzivc2021mixed} is unable to produce meaningful segmentation maps. When a number of anomalous training images are also segmented ($N>0$), the results improve even further, and significantly outperform the results of \cite{bovzivc2021mixed}. When using the full annotated training set of 246 segmented examples, DSR achieves anomaly detection performance near that of the fully supervised method~\cite{bovzivc2021mixed} and outperforms the recent fully supervised method presented in \cite{leiMachines2021}. Furthermore, DSR achieves the highest localization performance $AP_{loc}$ under all training settings, significantly outperforming~\cite{bovzivc2021mixed}. These results demonstrate that DSR can efficiently work in supervised settings as well, utilising all information available.

\begin{table}[hb]
\centering
\resizebox{0.40 \linewidth}{!}{\begin{tabular}{c c c c c c c}
\hline
 & Method & $N$: & 0 & 16 & 53 & 246\\ \hline
\multirow{3}{*}{$AP_{det}$} & \cite{bovzivc2021mixed} && 73.3 & 83.2 & 89.1 & \textbf{95.4} \\ 
 & DSR && \textbf{87.2} & \textbf{91.4} & \textbf{94.6} & 95.2 \\
 & \cite{leiMachines2021} && - & - & - & 93.3\\ \hline
\multirow{2}{*}{$AP_{loc}$} & \cite{bovzivc2021mixed} && 1.0 & 45.1 & 52.2 & 67.6 \\ 
 & DSR && \textbf{61.4} & \textbf{71.2} &\textbf{ 81.6} & \textbf{85.5} \\
\hline
\end{tabular}}
\caption{Anomaly detection and localization on the KSDD2 dataset in a supervised settings w.r.t. number of used anomalous training images $N$ with ground truth masks.
}
\label{tab:ksdd2ms}
\end{table}

\subsection{Experimental results on MVTec}

\begin{figure*}
\centering
  \includegraphics[width=1.0\linewidth]{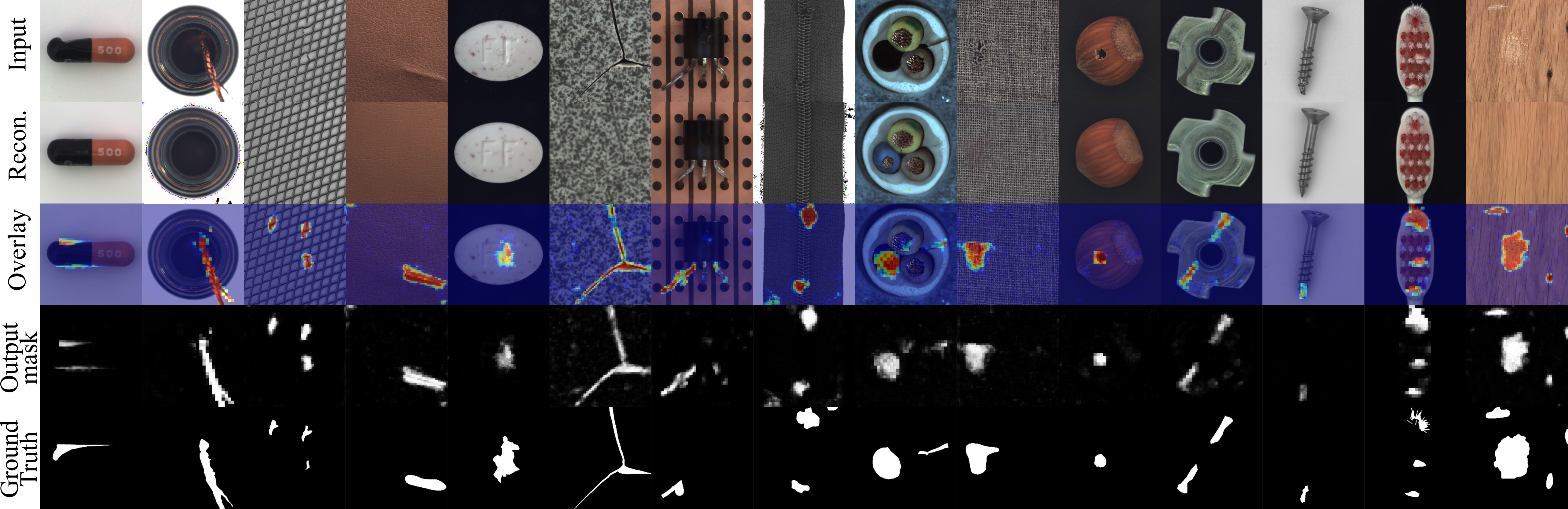}
\caption{Qualitative results on the MVTec dataset: the input images, outputs of the reconstruction network, input images with overlaid output masks, the output masks and the ground truth masks are shown in individual rows.}
\label{fig:mvtec_examples}
\end{figure*}

We also perform experiments on the MVTec anomaly detection dataset~\cite{bergmann2019mvtec}, which contains $15$ different texture and object classes, and has been established as the standard surface anomaly evaluation benchmark~\cite{li2021cutpaste,defard2021padim,zavrtanik2021draem,featspace2}.
The training set consists only of anomaly-free images, while the test set is comprised of anomalous as well as anomaly-free images.
The widely used AUROC metric is applied for image-level anomaly detection. 
Because only a fraction of the pixels in the test set are anomalous, the pixel-wise average-precision metric AP~\cite{zavrtanik2021draem} is used to evaluate the anomaly localization performance. It is more robust to class imbalance and better suited for anomaly localization evaluation than the commonly used pixel-wise AUROC.

\subsubsection{Comparison with the state-of-the-art}

\begin{table*}
\centering
\resizebox{1.0 \linewidth}{!}{\begin{tabular}{c c c c c c c c c c c c c c c c || c}
\hline
Method                        & bottle & capsule & grid  & leather & pill & tile & trans. & zipper & cable & carpet & hazelnut & m. nut & screw & toothbrush & wood & average \\ \hline \hline
\cite{bergmann2020uninformed} & 99.0   & 86.1    & 81.0  & 88.2    & 87.9 & 99.1 & 81.8       & 91.9   & 86.2  & 91.6   & 93.1     & 82.0      & 54.9  & 95.3       & 97.7 & 87.7 \\
\cite{zavrtanik2020riad}      & 99.9   & 88.4    & 99.6  & \textbf{100}     & 83.8 & 98.7 & 90.9       & 98.1   & 81.9  & 84.2   & 83.3     & 88.5      & 84.5  & \textbf{100}        & 93.0 & 91.7 \\
\cite{featspace2}              & \textbf{100}    & 92.3    & 92.9  & \textbf{100}     & 83.3 & 97.4 & 95.9       & 97.9   & \textbf{94.0}  & 95.5   & 98.7     & 93.1      & 81.2  & 95.8       & 97.6 & 94.4 \\
\cite{defard2021padim}        & 99.8   & 91.5    & 95.7  & \textbf{100}     & 94.4 & 97.4 & \textbf{97.8}       & 90.9   & 92.2  & 99.9   & 93.3     & 99.2      & 84.4  & 97.2       & 98.8 & 95.5 \\
\cite{li2021cutpaste}         & 98.2   & 98.2    & \textbf{100}   & \textbf{100}     & 94.9 & 94.6 & 96.1       & 99.9   & 81.2  & 93.9   & 98.3     & \textbf{99.9}      & 88.7  & 99.4       & \textbf{99.1} & 96.1 \\
\cite{zavrtanik2021draem}     & 99.2   & \textbf{98.5}    & 99.9  & \textbf{100}     & \textbf{98.9} & 99.6 & 93.1       & \textbf{100}    & 91.8  & 97.0   & \textbf{100}      & 98.7      & 93.9  & \textbf{100}        & \textbf{99.1} & 98.0 \\ 
\textbf{DSR}                          & \textbf{100}    & 98.1    & \textbf{100}   & \textbf{100}     & 97.5 & \textbf{100}  & \textbf{97.8}       & \textbf{100}    & 93.8  & \textbf{100}    & 95.6     & 98.5      & \textbf{96.2}  & 99.7       & 96.3 & \textbf{98.2} \\
\end{tabular}}
\caption{Results of anomaly detection on MVTec dataset (AUROC) with the average score over all classes ($avg$) in the last column.}
\label{tab:resSota}
\end{table*}

We evaluate DSR against recent state-of-the-art surface anomaly detection methods. The experimental results are presented in Tables~\ref{tab:resSota} and \ref{tab:resSotaLoc} for image-level anomaly detection and for pixel-level anomaly localization, respectively, and show that DSR outperforms the recent state-of-the-art approaches. It achieves an average AUROC of $98.2\%$, while maintaining a strong anomaly localization performance with an AP score of $70.2\%$. DSR outperforms the previous top performer in anomaly detection DR$AE$M~\cite{zavrtanik2021draem} on the mean AUROC score by $0.2$ percentage points (p.p.) and achieves superior performance on classes such as transistor, cable, carpet and screw, where near-distribution anomalies such as deformations are more prevalent. Qualitative results on the MVTec dataset are shown in Figure \ref{fig:mvtec_examples} and demonstrate that the detected anomalous regions resemble the ground truth anomaly maps to a high degree. 

\begin{table*}
\centering
\resizebox{1.0 \linewidth}{!}{\begin{tabular}{c c c c c c c c c c c c c c c c || c}
\hline
Method                        & bottle & capsule & grid  & leather & pill & tile & trans. & zipper & cable & carpet & hazelnut & m. nut & screw & toothbrush & wood & average \\ \hline \hline
\cite{bergmann2020uninformed} & 74.2   & 25.9    & 10.1  & 40.9    & 62.0 & 65.3 & 27.1       & 36.1   & 48.2  & 52.2   & 57.8     & 83.5      & 7.8   & 37.7       & 53.3 & 45.5 \\
\cite{zavrtanik2020riad}      & 76.4   & 38.2    & 36.4  & 49.1    & 51.6 & 52.6 & 39.2       & 63.4   & 24.4  & 61.4   & 33.8     & 64.3      & 43.9  & 50.6       & 38.2 & 48.2 \\
\cite{defard2021padim}        & 77.3   & 46.7    & 35.7  & 53.5    & 61.2 & 52.4 & \textbf{72.0}       & 58.2   & 45.4  & 60.7   & 61.1     & 77.4      & 21.7  & 54.7       & 46.3 & 55.0 \\
\cite{zavrtanik2021draem}     & 86.5   & 49.4    & 65.7  & \textbf{75.3}    & 48.5 & 92.3 & 50.7       & \textbf{81.5}   & 52.4  & 53.5   & \textbf{92.9}     & \textbf{96.3}      & \textbf{58.2}  & 44.7       & \textbf{77.7} & 68.4 \\ 
\textbf{DSR}                           & \textbf{91.5}   & \textbf{53.3}    & \textbf{68.0}  & 62.5    & \textbf{65.7} & \textbf{93.9} & 41.1       & 78.5   & \textbf{70.4}  & \textbf{78.2}   & 87.3     & 67.5      & 52.5  & \textbf{74.2}       & 68.4 & \textbf{70.2} \\
\end{tabular}}
\caption{Results (AP) of anomaly localization on MVTec dataset.}
\label{tab:resSotaLoc}
\end{table*}

\begin{figure}
\centering
  \includegraphics[width=0.7\linewidth]{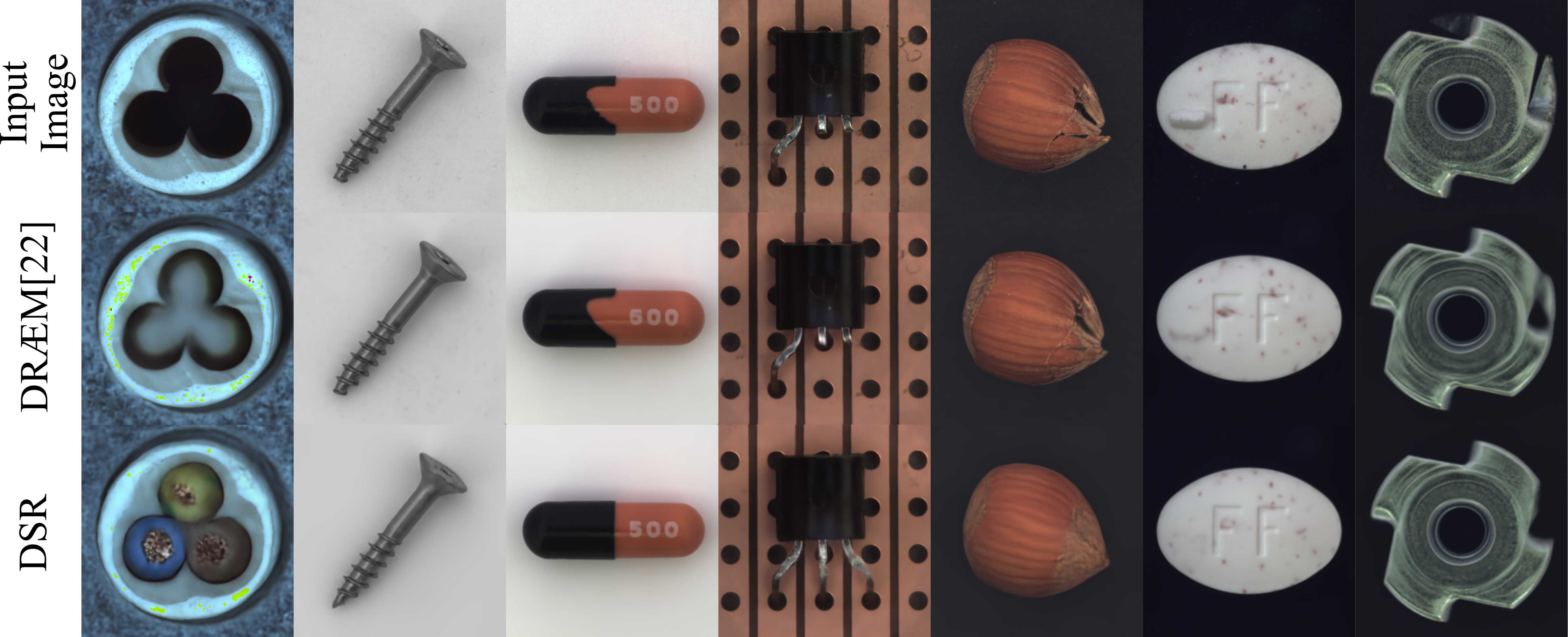}
\caption{
Qualitative reconstruction results. DSR far better reconstructs the anomalies by their corresponding in-distribution appearance than the state-of-the-art 
DR$\AE$M~\cite{zavrtanik2021draem}.
}
\label{fig:recon_ex}
\end{figure}

Further qualitative comparison is shown in Figure~\ref{fig:recon_ex}. Both DSR and DR$\AE$M~\cite{zavrtanik2021draem} are trained to restore the normality of images corrupted by simulated anomalies. However, DR$\AE$M~\cite{zavrtanik2021draem} generates the anomalies in the image space from an out-of-distribution dataset, while DSR generates the anomalies within the quantized feature space, making it more difficult for the reconstruction network to overfit to the simulated anomaly appearance. This results in DSR having a more robust image normality restoration capability that is insensitive to near-in-distribution anomalies such as deformations. Figure~\ref{fig:recon_ex} compares the reconstruction results of both methods. Note that DSR produces a more realistic reconstruction. Due to the reconstruction network not overfitting to the synthetic anomaly appearance, DSR can recognize deformations as deviations from normality and reconstruct them accordingly.

\subsection{Ablation Study}

Additional experiments on the MVTec dataset provide further insights into DSR. 

\textbf{Anomaly source evaluation.} 
We evaluate the impact of training using our feature-based anomalies and compare it to the training with image-based anomalies generated from out-of-distribution datasets as proposed in~\cite{zavrtanik2021draem}. The results are shown in Table~\ref{tab:ablation}(a), where DSR$_{img}$ denotes the results when using image-based anomalies during training. We can see that feature-based anomaly generation (DSR) outperforms the image-based approach in anomaly localization and detection.

\textbf{Anomaly feature sampling.} 
To generate simulated anomalies, DSR samples vectors from the codebook $K$ according to a designed process presented in Section~\ref{sec:anomgen}. We compare this approach to one that samples the codebook with a uniform probability for each vector. The results (DSR$_{random}$) presented in Table \ref{tab:ablation}(b) show a significant drop in performance. The sampling method, therefore, plays an important role. The proposed approach generates anomalous regions from vectors that are close to the extracted vectors generating simulated near-distribution anomalies that leads to superior results in terms of anomaly detection as well as localization. DSR is robust to the choice of the similarity bound parameter, retaining a good anomaly detection performance for a wide range of $\lambda_s$ values (Table \ref{tab:simBound}).
\begin{table}
\centering
\begin{tabular}{c |  c c c c c c }\hline
 $\lambda_s$ & 0.01 & 0.02 & 0.05 & 0.1 & 0.2 & 0.5 \\ \hline
 AUROC$_{det}$ & 97.5 & 98.0 & 98.2 & 97.7 & 97.1 & 95.5 \\
\end{tabular}
\caption{Results on MVTec using various similarity bound $\lambda_s$ values. 
}
\label{tab:simBound}
\end{table}

\textbf{Reconstruction loss components.} 
Table \ref{tab:ablation}(c) shows the impact of individual reconstruction loss components of Eq.~(\ref{eq:reconloss}), where the reconstruction is conditioned both on the feature $L_{feat}$ and image $L_{img}$ reconstruction losses, which are the second and the third components in Eq.~(\ref{eq:reconloss}), respectively. During training DSR$_{L_{img}}$ uses only the image reconstruction loss and DSR$_{L_{feat}}$ uses only the feature reconstruction loss. There is a significant loss in performance when using only individual loss components. Relying on both image and feature reconstruction losses during training results in a more robust normality reconstruction model leading to a $2$ p.p. higher average image-level AUROC as well as significantly higher localization AP scores.

\textbf{Upsampling module}
Table \ref{tab:ablation}(d) shows the effect of removing the Upsampling module of the network, leading to a drop in localization performance. The anomaly detection performance remains the same as the image-level score is extracted from the lower resolution output anomaly map $M$.

\begin{table}
\centering
\resizebox{0.8 \linewidth}{!}{\begin{tabular}{c | c | c | c | c c | c }
\hline
   \multicolumn{2}{c}{ } & \multicolumn{1}{c}{(a)} & \multicolumn{1}{c}{(b)} & \multicolumn{2}{c}{(c)} & \multicolumn{1}{c}{(d)} \\ \hline
 Experiment & DSR & DSR$_{img}$ &  DSR$_{random}$ & DSR$_{L_{img}}$ & DSR$_{L_{feat}}$ & DSR$_{U-}$ \\ \hline
 AUROC$_{det}$ & 98.2 & 97.8 & 95.6 & 96.3 & 95.2 & 98.2 \\
 AP$_{loc}$ & 70.2 & 67.5 & 62.5 & 67.0 & 61.8 & 65.2 \\ \hline

\end{tabular}}
\caption{Ablation study results on MVTec: (a) using out-of-distribution texture-based anomalies (DSR$_{img}$) in training; (b) unconstrained uniform anomaly sampling (DSR$_{random}$); (c) training with only image reconstruction loss (DSR$_{L_{img}}$) and with only the feature reconstruction loss (DSR$_{L_{feat}}$); (d) Performance without the Upsampling module (DSR$_{U-}$).
}
\label{tab:ablation}
\end{table}

\section{Conclusion}

We proposed DSR, a discriminative surface anomaly detection method based on dual image reconstruction branch architecture with discretized latent representation. Such representation allows controlled generation of synthetic anomalies in feature space, which, in contrast to the state-of-the-art methods that generate anomalies in image space, makes no assumption about the anomaly appearance and does not rely on image-level heuristics. Our anomaly generation approach produces near-in-distribution synthetic anomalies resulting in more robustly trained reconstruction capabilities and detection of anomalies whose appearance is close to the normal appearance. 

On the recent challenging real-world KSDD2 dataset~\cite{bovzivc2021mixed}, it outperforms all other unsupervised surface anomaly detection methods by $10\%$ AP in anomaly detection and $35\%$ AP in anomaly localization tasks. DSR also significantly outperforms the weakly-supervised model presented in \cite{bovzivc2021mixed}. Moreover, we showed that the proposed approach can be extended to utilise also pixel-wise annotated  anomalous training samples. When used in such supervised settings, it considerably improves the results of related supervised methods \cite{bovzivc2021mixed} when only a few annotated examples are available. This demonstrates the potential of DSR to be used in real-world settings where the number of available anomalous images is typically too low to train fully supervised methods. DSR also achieves state-of-the-art results on the standard MVTec anomaly detection dataset~\cite{bergmann2019mvtec}.

The ablation study shows that the sampling method used for anomaly generation impacts the final anomaly detection results significantly, which suggests that a more complex feature sampling method may improve results further, which we leave for our future research endeavours.

In addition to it's good performance across several tasks and it's ability to utilize anomalous training samples that are available in some practical scenarios, DSR is fast, running at $58$ FPS, making it a good choice for real-world industrial applications with real-time requirements. \\

\noindent \textbf{Acknowledgement} \textit{This work was supported by Slovenian research agency programs J2-2506, L2-3169, P2-0214. Vitjan Zavrtanik was supported by the Young researcher program of the ARRS. }

{\small
\bibliographystyle{splncs04}
\bibliography{dsr}
}

\end{document}